\documentclass[11pt]{article}

\usepackage[preprint]{acl}

\usepackage{times}
\usepackage{latexsym}
\usepackage{latexsym, amsmath, amssymb, amsfonts, mathrsfs}

\usepackage[T1]{fontenc}

\usepackage[utf8]{inputenc}

\usepackage{microtype}

\usepackage{inconsolata}

\usepackage{graphicx}

\usepackage{booktabs}          
\usepackage{pgfplots}          
\pgfplotsset{compat=1.18}
\usetikzlibrary{positioning, arrows.meta, calc}

\usepackage{multicol}                 

\title{Discourse-Aware Policy Analysis with Argumentation: A Hybrid LLM–Symbolic Framework for Disaster Governance}

\author{Stylianos Loukas Vasileiou \\
  New Mexico State University / Address line 1 \\
  Affiliation / Address line 2 \\
  Affiliation / Address line 3 \\
  \texttt{email@domain} \\\And
  Second Author \\
  Affiliation / Address line 1 \\
  Affiliation / Address line 2 \\
  Affiliation / Address line 3 \\
  \texttt{email@domain} \\}

\author{
 \textbf{Stylianos Loukas Vasileiou\textsuperscript{1}},
 \textbf{Olga Derendiaeva\textsuperscript{2}}
\\
 \textsuperscript{1}New Mexico State University,
 \textsuperscript{2}Sun Yat-Sen University,
}

\begin{document}
\maketitle
\begin{abstract}
Policy documents shape governance outcomes, but their reasoning is often implicit. Participatory commitments and managerial control routinely coexist in the same text, and the tensions between them are rarely stated directly. Existing computational approaches to policy discourse cannot express the \emph{frame-mediated} relations that drive these tensions, where one argument narrows or instrumentalizes another rather than rejecting it. End-to-end summarization by large language models produces fluent text but offers little structure that domain experts can inspect or contest. We present \textsc{Apaf}, a hybrid LLM--symbolic pipeline that operationalizes critical discourse analysis as a quantitative bipolar argumentation framework over policy text. Arguments are first classified into \emph{deliberative} or \emph{managerial} frames. Four frame-mediated relation subtypes (\emph{agency reduction}, \emph{agenda shift}, \emph{instrumental support}, and \emph{normative support}) are then produced by deterministic rules over LLM-extracted features. We release a novel dataset of 100 sub-documents of disaster-risk-reduction policy from the USA, UK, Canada, and Australia, and show that the resulting argument graphs are accurate, interpretable, and stable across jurisdictions.
\end{abstract}

\section{Introduction}
\label{sec:intro}

\begin{figure*}[!t]
\centering
\includegraphics[width=\textwidth]{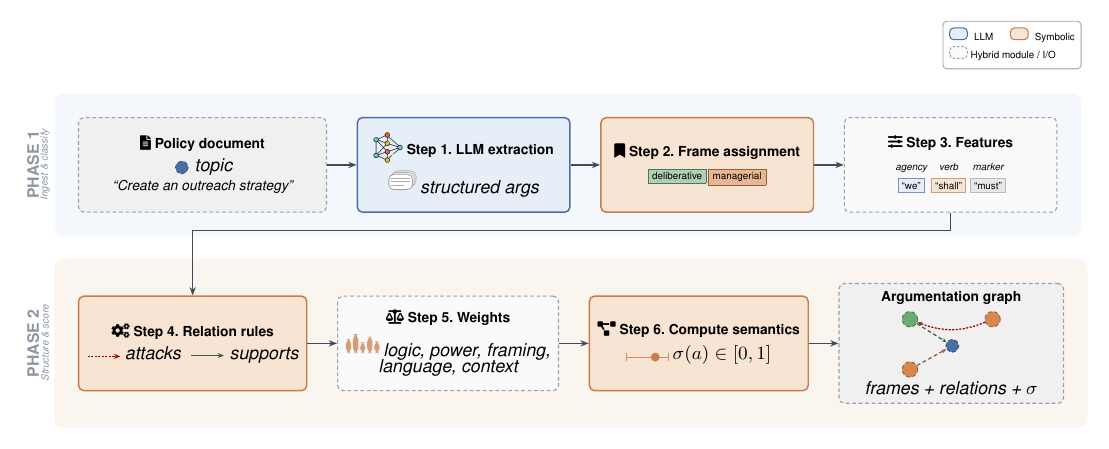}
\caption{\textbf{The \textsc{Apaf} pipeline.} Six steps convert a policy document and a topic argument into a quantitative bipolar argumentation graph. \textbf{Phase~1:} an LLM extracts structured arguments (Step~1), a symbolic classifier assigns each one a \emph{deliberative} or \emph{managerial} frame (Step~2), and a hybrid feature detector picks up agency markers, verb classes, and instrumental cues (Step~3). \textbf{Phase~2:} symbolic rules fire frame-mediated attack and support relations (Step~4), each argument receives a base score across five linguistic dimensions (Step~5), and DF-QuAD gradual semantics propagates these scores through the graph to a final strength $\sigma(a)\in[0,1]$ (Step~6). The output (right) is an argumentation graph with a topic node (blue) and \emph{deliberative} (green) and \emph{managerial} (orange) argument nodes, related by support (solid) and attack (dashed) edges. Every relation in the output is traceable to either a recorded LLM call or an explicit rule, so the resulting graph is inspectable and contestable.}
\label{fig:pipeline}
\end{figure*}

Decisions about who is exposed to risk, who decides how risk is managed, and whose voice counts in the planning process are increasingly mediated by long policy documents. In disaster governance, successive international and national frameworks have introduced participatory and deliberative language into systems that remain, in practice, top-down and compliance-driven \citep{sylves2008disaster}. The gap between ``whole-community engagement'' as a normative commitment and the same phrase as a checkbox for federal funding is rarely stated directly. Instead, it is expressed through how arguments are framed and which actors are assigned agency. Understanding governance therefore requires understanding the implicit reasoning of policy text, at scale and across the jurisdictions in which policy is produced.

Two limitations stand in the way. First, computational argument mining \citep{lawrence2019argument,cabrio2018five,lippi2016argumentation} has matured on essays, debates, and online discussions, where attacks tend to be explicit. Its standard ontology of support versus direct attack cannot express the \emph{frame-mediated} relations that drive real policy discourse. In such relations, one argument does not reject another. Instead, it narrows the other argument, conditions it on compliance, or replaces a shared-agency actor with a unilateral one. Recent work on implicit \citep{saadatyazdi2023uncovering} and scheme-based \citep{goffredo2023argument} relations moves in this direction, but no existing system models the specific frame-mediated dynamics that drive governance discourse. Second, LLMs offer fluent extraction \citep{ziems2024can,dagdelen2024structured} but yield little structure that a domain expert can inspect, modify, or contest \citep{rudin2019stop,jacovi2020faithfulness,lazar2024llms}. Stakeholders in high-stakes governance need representations that are not just plausible but also auditable \citep{vasileiou2026argumentative}.

We address both limitations with \textsc{Apaf} (Adaptive Policy Argumentation Framework), a hybrid LLM--symbolic pipeline that operationalizes critical discourse analysis (CDA) \citep{fairclough2003analysing,habermas1984theory} as a Quantitative Bipolar Argumentation Framework (QBAF) \citep{rago2016discontinuity,baroni2018how} over policy text. An LLM first extracts arguments from a document and classifies them into one of two governance frames, \textit{deliberative} or \textit{managerial}~\citep{dryzek2002deliberative}. Deterministic symbolic rules then produce four frame-mediated relation subtypes that are not modeled in classical argument mining. These are \emph{agency reduction} (managerial-tokenistic claims that displace deliberative-empowerment ones), \emph{agenda shift} (unilateral actors that displace shared-symmetrical ones), and the corresponding \emph{instrumental} and \emph{normative} support relations. The full argumentation graph is then solved by gradual semantics \citep{rago2016discontinuity} to yield final argument strengths. Because the pipeline interleaves LLM extraction with symbolic classification and rule firing, every relation and weight is traceable to either an explicit rule or a recorded LLM call. This design follows the recent line of faithfulness-by-construction systems \citep{lyu2023faithful,xu2024symcot,freedman2025argumentative}.

We evaluate \textsc{Apaf} on a new dataset of 100 sub-documents of disaster-risk-reduction policy from four anglophone democracies (USA, UK, Canada, Australia), annotated by two researchers using a CDA-grounded codebook over two coding passes. We evaluate argument extraction end-to-end, and we evaluate frame assignment and relation classification on annotated arguments fed in directly, isolating downstream quality from extraction errors. We score relations at three nested levels of granularity (\emph{Detection}, \emph{Polarity}, and \emph{Subtype}), defined in \S\ref{sec:experiments}.

Our contributions are as follows:
\begin{itemize}
\setlength\itemsep{0.1em}
\item A \textbf{discourse-aware argumentation framework} that introduces frame-mediated attack and support subtypes (agency reduction, agenda shift, instrumental and normative support) grounded in critical discourse analysis.
\item A \textbf{hybrid LLM--symbolic pipeline} that builds argumentation graphs from policy text with traceable, contestable outputs, and an \textbf{empirical evaluation} demonstrating its efficacy.
\item A \textbf{multi-country governance corpus} of 100 sub-documents from USA, UK, Canada, and Australia, annotated for arguments, frames, and typed relations.
\end{itemize}

\section{Related Work}
\label{sec:related}

\paragraph{Argument mining for policy and political text.}
Argument mining has matured on persuasive essays \citep{stab2014identifying,stab2017parsing}, web discourse \citep{habernal2017argumentation,chakrabarty2019ampersand}, and end-to-end neural pipelines \citep{eger2017neural,morio2022end}. Recent surveys cover the field in depth \citep{lawrence2019argument,cabrio2018five,lippi2016argumentation}. Closer to governance text, work on political debates such as the 2016 US presidential debates \citep{visser2019argumentation,haddadan2019disputool,goffredo2023disputool} and on fallacy detection in political discourse \citep{goffredo2023argument,pan2024llms} models explicit argumentative moves. Legal argument mining \citep{habernal2023mining,xu2022multi,lippi2019claudette} extracts structured claims from court decisions and contractual text. Our setting differs in two ways. First, governance discourse rarely contains direct rebuttals. Attacks are produced through frame shifts and changes in who is assigned agency, and existing typologies do not represent these moves. Recent work on implicit \citep{saadatyazdi2023uncovering} and quality-aware \citep{wachsmuth2017computational} argument mining moves toward this gap but does not close it. Second, we evaluate across four national jurisdictions, while prior policy argument mining is overwhelmingly US-centric.

\paragraph{LLMs for structured extraction, and hybrid neurosymbolic pipelines.}
Schema-bound LLM extraction is now a strong baseline across high-stakes domains \citep{dagdelen2024structured,han2023empirical,ghosh2024reliable,shrimal2025parse,ziems2024can}, although format restrictions can degrade reasoning quality \citep{tam2024format} and structured-output interfaces remain a usability bottleneck for practitioners \citep{liu2024structured}. To recover interpretability, a growing body of work splits the workload between a neural front-end and a symbolic back-end. Faithful chain-of-thought translates language into executable programs \citep{lyu2023faithful}, symbolic chain-of-thought decomposes reasoning into solver-checkable steps \citep{xu2024symcot}, and \citet{allen2025sound} ground formal interpretations in LLM outputs. The broader agenda is reviewed in \citet{hamilton2022neurosymbolic}. Within argument mining specifically, \citet{vandermeer2024hybrid} and \citet{li2025large} survey hybrid intelligence designs and the rise of LLM-based pipelines. \textsc{Apaf} follows this hybrid pattern. It calls the LLM only where ambiguous classification benefits from world knowledge (argument extraction, agency typing, contradiction counting) and delegates relation construction and gradual-strength computation to deterministic components.

\paragraph{Frames, power, and agency in computational social science.}
Computational framing analysis is anchored by the Media Frames Corpus \citep{card2015media,card2016analyzing} and its successors \citep{mendelsohn2021modeling,frermann2023conflicts,otmakhova2024media}, with extensions to agenda setting \citep{tsur2015frame,field2018framing}, cross-country generalization \citep{daffara2025generalizability}, and morality-laden targeting \citep{roy2021identifying,khanehzar2021framing}. Power and agency in text are operationalized by the connotation-frame line \citep{rashkin2016connotation,sap2017connotation,ma2020powertransformer,antoniak2023riveter,khanehzar2023probing}, with applications to dehumanization \citep{mendelsohn2020framework}, narrative dynamics \citep{antoniak2019narrative,piper2021narrative}, and causal micro-narratives \citep{heddaya2024causal}. \textsc{Apaf} differs from these lines in two ways. First, frames are not topic categories but \emph{governance rationalities} (deliberative or managerial) that license specific relation types between arguments. Second, agency and power appear as argument-level features rather than document- or entity-level descriptors, and when combined with frames they deterministically yield typed attack and support relations in the argumentation graph.

\paragraph{Quantitative argumentation and contestable AI.}
QBAFs \citep{baroni2018how,amgoud2018weighted,amgoud2017acceptability} extend Dung-style argumentation \citep{dung1995acceptability} with weighted bipolar structure. We use the DF-QuAD gradual semantics of \citet{rago2016discontinuity}, with recent modular variants in \citet{rago2025methodology}. DF-QuAD has been applied to design decisions \citep{baroni2015automatic}, forecasting \citep{irwin2022forecasting}, and as a bridge to neural networks \citep{potyka2021interpreting}. Argumentation has been argued to be the appropriate substrate for contestable AI \citep{cyras2021argumentative,vassiliades2021argumentation,leofante2024contestable,freedman2025argumentative,castagna2024computational,cocarascu2019extracting,vasileiou2026argumentative}. This complements the broader case for inherently interpretable models in high-stakes settings \citep{rudin2019stop,rudin2022principles,jacovi2020faithfulness,lyu2024faithful,zhao2024xaillm,miller2019explanation}. To our knowledge, \textsc{Apaf} is the first instantiation of this argumentation-for-XAI agenda over real governance policy at the corpus scale.

\section{Theoretical Background}
\label{sec:background}

\paragraph{Governance rationalities and frames.}
We treat policy documents as sites of interaction between competing \emph{governance rationalities} \citep{habermas1984theory,dryzek2002deliberative}. A \emph{communicative} rationality is oriented toward participation, deliberation, and the inclusion of plural voices. An \emph{instrumental} rationality is oriented toward administration, compliance, and measurable implementation. In policy practice the two rationalities coexist within the same document. Participatory narratives supply normative legitimacy, while managerial reasoning supplies procedural legitimacy through regulation and accountability \citep{scharpf1999governing}. Following deliberative-governance theory \citep{dryzek2002deliberative}, we operationalize these tensions through two discourse \emph{frames}. The \emph{deliberative} frame covers participation, co-creation, and shared agency. The \emph{managerial} frame covers compliance, institutional authority, and procedural implementation. The two frames are not purely oppositional. Policy text routinely contains hybrid structures in which managerial framing supports deliberative goals instrumentally, or in which participatory concepts narrow into tokenistic requirements \citep{arnstein1969ladder,fairclough2003analysing}. These hybrid structures are precisely what our frame-mediated relation typology aims to capture.

\paragraph{QBAFs and DF-QuAD.}
A QBAF \citep{baroni2018how,amgoud2018weighted} is a tuple $\langle \mathcal{A}, \tau, \mathcal{R}^-, \mathcal{R}^+ \rangle$ of arguments $\mathcal{A}$, a base-score function $\tau:\mathcal{A} \to [0,1]$, an attack relation $\mathcal{R}^- \subseteq \mathcal{A} \times \mathcal{A}$, and a support relation $\mathcal{R}^+ \subseteq \mathcal{A} \times \mathcal{A}$. A gradual semantics assigns a final strength $\sigma(a) \in [0,1]$ to each argument by propagating influence along edges. We use DF-QuAD \citep{rago2016discontinuity}, which aggregates attackers and supporters with a discontinuity-free product operator and combines the two via a conservative influence function.

\section{The \textsc{Apaf} Framework}
\label{sec:framework}

\textsc{Apaf} converts a policy document into a QBAF in six steps (Figure~\ref{fig:pipeline}). Step~1 extracts arguments from the text. Step~2 assigns a frame to each argument. Step~3 detects discourse features (agency, verbs, and instrumental markers). Step~4 produces typed relations through symbolic rules. Step~5 computes a base score for each argument across five linguistic dimensions. Step~6 propagates these scores through the relation graph using DF-QuAD gradual semantics. For each document we also instantiate a fixed \emph{topic} argument $a_0$ that represents the document's stated objective, and we add topic-relative relations through dedicated rules described below.

\subsection{Arguments and frames}
\label{sec:args-frames}

\paragraph{Argument extraction.}
Given a document $D$, we segment $D$ into overlapping chunks and prompt an LLM with a schema-constrained prompt \citep{liu2024structured} to return explicit arguments. An argument must satisfy four criteria, namely assertive content, an identifiable subject, a normative or procedural orientation, and policy relevance. The full extraction prompt can be found in Appendix~\ref{app:prompts}. 
Each extracted argument $a$ is stored as a tuple $(\textsf{id}_a, \textsf{text}_a, \textsf{span}_a)$ that carries an extraction-mode tag.

\paragraph{Frame assignment.}
Each argument receives a frame $\phi(a) \in \{\textsf{deliberative}, \textsf{managerial}\}$ following the typology of \S\ref{sec:background}. Frames are computed in two stages. First, a keyword scan over a CDA-curated lexicon (Appendix~\ref{app:kw-frames})
returns a score for each frame. Second, a noun--verb matcher triggers when a managerial verb governs a deliberative noun, or vice versa, and can override the keyword scan. Importantly, the topic argument $a_0$ is always assigned the $\textsf{deliberative}$ frame by convention.

\subsection{Discourse features}
\label{sec:features}

For each argument $a$ we detect three feature families that the relation rules consume.

\begin{enumerate}\setlength\itemsep{0.2em}
\item \textbf{Agency} $\alpha(a)$ classifies who is identified as the actor. We use six types. Three are managerial, namely \textsf{unilateral\_top} (\emph{FEMA, the Federal government}), \textsf{unilateral\_mid} (\emph{the State, the jurisdiction, you}), and \textsf{directed} (\emph{the planning team, the department}). Two are deliberative, namely \textsf{shared\_symmetrical} (\emph{the whole community, the public}) and \textsf{horizontal} (\emph{mutual, reciprocal, two-way}). The sixth type is \textsf{none}, used when no actor is identifiable. Detection combines a regex scan with an LLM classifier that is invoked only when the regex is ambiguous.
\item \textbf{Verb class} $\nu(a)$ flags \emph{tokenistic} verbs such as \emph{educate}, \emph{inform}, and \emph{invite}, and \emph{empowerment} verbs such as \emph{co-decide}, \emph{share}, and \emph{deliberate}. This follows the spirit of connotation-frame lexica \citep{sap2017connotation,antoniak2023riveter} and the Ladder of Citizen Participation of \citet{arnstein1969ladder}.
\item \textbf{Instrumental markers} $\iota(a) \in \{\textsf{delib}, \textsf{mgr}, \emptyset\}$ flag whether the argument uses instrumental language oriented toward a deliberative or managerial goal. For instance, \emph{``to obtain federal funding''} is a managerial marker, and \emph{``to enable shared decision-making''} is a deliberative marker.
\end{enumerate}

Note that an argument can carry both a managerial and a deliberative signal.

\subsection{Frame-mediated relation rules}
\label{sec:relations}

The core mechanism of \textsc{Apaf} is a typed set of \emph{frame-mediated} relations that fire deterministically over argument pairs $(a_s, a_t)$. At most one relation is produced per directed pair. Each subtype is described below at the conceptual level. More details are given in Appendix~\ref{app:rules} and Appendix~\ref{app:keywords}.

\paragraph{Attack rules.}
Two attack subtypes capture the implicit conflicts of governance discourse motivated in \S\ref{sec:intro}. \emph{Agency reduction} fires when a managerial-framed argument with tokenistic verbs is paired with a deliberative-framed argument with empowerment verbs. The managerial argument narrows the deliberative one by replacing empowerment with passive participation. \emph{Agenda shift} fires when a managerial-framed argument whose actor is unilateral or directed is paired with a deliberative-framed argument whose actor is shared-symmetrical or horizontal. The managerial argument reassigns agency away from the community. The two subtypes are conceptually distinct but frequently co-occur. A paragraph that empowers a state agency \emph{to educate the public} typically realizes both at once.

\paragraph{Support rules.}
Two support subtypes distinguish \emph{instrumental} from \emph{normative} support, a distinction that classical argument mining collapses but that is central to deliberative-governance theory \citep{dryzek2002deliberative}. An instrumentally-supporting argument advances a target as a means to compliance or implementation. A normatively-supporting argument advances it as a shared end. \textsc{Apaf} produces instrumental support either when a managerial argument carries an instrumental marker for the deliberative goal of a deliberative target, or when a managerial-to-managerial pair carries a managerial-side instrumental marker and the two arguments are close in sentence-embedding space. Normative support fires between two deliberative arguments that are similarly close in embedding space. 

\paragraph{Topic rules.}
Two final rules connect every framed argument to the document's topic argument $a_0$. A managerial argument supports $a_0$ instrumentally, while a deliberative argument supports $a_0$ normatively. These rules play a central role in the gradual-semantics output, as they propagate every argument's evidence toward the topic.

\subsection{Weights and gradual semantics}
\label{sec:weights}

Each argument $a$ receives an initial base score $\tau(a) \in [0,1]$ defined as
\begin{equation}
\tau(a) = \frac{1}{5} \sum_{d \in \mathcal{D}} w_d(a),
\label{eq:tau}
\end{equation}
where $\mathcal{D} = \{\text{logic}, \text{power}, \text{framing}, \text{language},$ \\ $\text{context}\}$ and each $w_d(a) \in [0,1]$. Specifically,

\begin{itemize}
\setlength\itemsep{0.001em}
    \item \textbf{Logic} is a decreasing function of the LLM-counted contradictions involving $a$, mapping zero contradictions to $1.0$, one to $0.7$, and two or more to $0.3$.

    \item \textbf{Power} is a frame-conditioned lookup over the agency type $\alpha(a)$ \citep{gordon1980power}.

    \item \textbf{Framing} is the density of frame-consistent keywords in $a$ \citep{fairclough2013political}.

    \item \textbf{Language} maps the strongest modal verb in $a$ to $1.0$, $0.5$, or $0.3$ for strong, weak, or absent modality \citep{fairclough2013political}.

    \item \textbf{Context} scores institutional-lexicon alignment with known policy documents such as the Sendai Framework \citep{undrr2015sendai} and FEMA.

\end{itemize}

Note that the topic argument $a_0$ receives a fixed base score $\tau(a_0)$ that we treat as a hyperparameter, set to $0.5$ in our experiments.

Finally, given $\mathcal{A}$, $\tau$, $\mathcal{R}^-$, and $\mathcal{R}^+$, we apply DF-QuAD \citep{rago2016discontinuity} to compute a final strength $\sigma(a) \in [0,1]$ for every argument. The output is the complete QBAF $\langle \mathcal{A}, \tau, \mathcal{R}^-, \mathcal{R}^+, \sigma \rangle$, together with the rule or LLM trace that produced each argument and relation. We render this output in an interactive graph explorer that lets a domain expert edit any frame, relation, or score and re-solve the graph in a single pass; the interface is described in Appendix~\ref{app:interface}\label{sec:hitl}.

\section{Dataset}
\label{sec:dataset}

\paragraph{Source materials.}
We curated a multi-country corpus of disaster-risk-reduction policy from four anglophone democracies, namely the United States, the United Kingdom, Canada, and Australia. The four countries share a working language, comparable institutional structures (developed federal or devolved democracies), and active engagement with the UN Sendai Framework for Disaster Risk Reduction \citep{undrr2015sendai}. They also span different agencies, including FEMA in the US, the Civil Contingencies Secretariat and Local Resilience Forums in the UK, Public Safety Canada, and the National Emergency Management Agency in Australia. This combination provides a controlled setting for cross-country comparison while keeping institutional variation explicit. Representative source documents include the FEMA Local Mitigation Planning Handbook \citep{fema2025handbook} and the National Mitigation Framework \citep{fema2016framework}, together with city-level plans such as the La Ca{\~n}ada Flintridge Local Hazard Mitigation Plan \citep{lacanada2024lhmp} for the US. For the UK they include the Defra guide for local councils \citep{defra2010adapting} and the Redbridge Local Flood Risk Management Strategy \citep{redbridge2024flood}. For Canada they include the Public Safety Canada Emergency Management Strategy for Canada \citep{psc2019emstrategy}. For Australia they include the Council of Australian Governments National Strategy for Disaster Resilience \citep{coag2011nsdr}. From each country we segmented 25 sub-documents, for a total of 100 sub-documents.

\paragraph{Annotation protocol.}
Two researchers with expertise in policy analysis and argumentation jointly developed the annotation codebook over multiple iterations. The codebook is grounded in Fairclough's three-dimensional model of critical discourse analysis \citep{fairclough1992discourse,fairclough2003analysing} and operationalized through the typology of \S\ref{sec:background} and \S\ref{sec:framework}. It specifies four annotation layers per sub-document, covering argument spans (using the four-criterion definition of \S\ref{sec:args-frames}), governance frames (\textsf{deliberative} or \textsf{managerial}), typed support and attack relations, and a topic argument $a_0$ stating the sub-document's objective. 

We used two methodological safeguards to control annotator bias. First, the two annotators jointly developed the codebook on a shared pilot subset and reconciled all disagreements through discussion before the main coding pass. Second, the primary annotator re-coded the full corpus to check intra-annotator stability, with cases where the two passes disagreed resolved by joint discussion with the second annotator. 
Table~\ref{tab:corpus} summarizes the corpus. Across all four countries, instrumental support relations dominate over normative support, which is an empirical reflection of the managerial framing layered atop deliberative language that motivates the framework. The corpus, the codebook, the per-layer documentation, and the full pipeline code will be released to the community upon acceptance.

\begin{table}[t]
\centering
\begin{tabular}{@{}l|rrr@{}}
\toprule
\textbf{Country} & \textbf{Args} & \textbf{Rels} & \textbf{Rel/Arg} \\
\midrule
USA       & 427  & 3392 & 7.9 \\
UK        & 242  &  689 & 2.8 \\
Canada    & 274  & 1132 & 4.1 \\
Australia & 336  & 2002 & 6.0 \\
\midrule
\textbf{Total} & \textbf{1279} & \textbf{7215} & \textbf{5.6} \\
\bottomrule
\end{tabular}
\caption{Dataset statistics.}
\label{tab:corpus}
\end{table}

\section{Experiments}
\label{sec:experiments}

We now focus on evaluating our framework on extraction and relation classification. A human user study on the efficacy and usefulness of the interactive interface is left to future work.

\paragraph{Setup.}
We evaluate \textsc{Apaf} on the dataset described in \S\ref{sec:dataset}. In our configuration, we used \texttt{gpt-5.4} \citep{openai2023gpt4} as the downstream LLM. Extraction is evaluated end-to-end, with the pipeline running from raw text and predicted arguments compared against the annotated set via embedding-based alignment. Frame assignment and relation classification are evaluated on annotated arguments fed in directly, so that downstream metrics are not confounded by extraction errors. All numbers below are averaged over 3 runs, with ${\pm}$ indicating one standard deviation across runs.

\paragraph{Metrics.}
For \emph{extraction}, we report P/R/$F_1$ of the alignment between predicted and annotated arguments. For \emph{frames}, we report accuracy over non-topic arguments. For \emph{relations}, we report P/R/$F_1$ at three nesting levels, \textit{Detection}, \textit{Polarity}, and \textit{Subtype}. Detection requires only that the directed pair exists, Polarity additionally requires the correct support or attack label, and Subtype additionally requires the correct frame-mediated subtype (agency-reduction, agenda-shift, instrumental-support, or normative-support). All P/R/$F_1$ numbers in this section are micro-averaged across sub-documents, following standard practice in relation extraction and argument mining \citep{stab2017parsing,morio2022end}. 

\paragraph{Results.}
Table~\ref{tab:overall} reports overall performance over all 100 sub-documents. The pipeline reaches $F_1 = 0.91$ on argument extraction and accuracy of $0.86$ on frame assignment over the 1279 annotated arguments, and a Detection $F_1$ of $0.73$ over 7,215 annotated relations. Adding the Polarity requirement drops $F_1$ to $0.66$, and adding the Subtype requirement drops it further to $0.58$, for an aggregate $15$-point loss across the two layers. The relative monotonicity matters. Roughly $10\%$ of correctly-identified directed pairs are mis-labeled at the Polarity level, and a further ${\sim}11\%$ of Polarity-correct pairs are mis-labeled at the Subtype level. The Subtype level is where our frame-mediated typology lives, and is therefore the level most exposed to feature errors in LLM-assisted agency typing and lexicon-based verb classification.

\begin{table}[t]
\centering
\small
\begin{tabular}{@{}ll|ccc@{}}
\toprule
\textbf{Block} & \textbf{Level} & \textbf{P} & \textbf{R} & \textbf{$F_1$/Acc} \\
\midrule
extraction & \textendash & .958$_{\pm.002}$ & .864$_{\pm.003}$ & .909$_{\pm.001}$ \\
\midrule
frames & \textendash & \textendash & \textendash & .857$_{\pm.002}$ \\
\midrule
relations & Detection & .703$_{\pm.001}$ & .751$_{\pm.004}$ & .726$_{\pm.003}$ \\
relations & Polarity & .635$_{\pm.001}$ & .678$_{\pm.002}$ & .656$_{\pm.001}$ \\
relations & Subtype & .563$_{\pm.003}$ & .601$_{\pm.001}$ & .581$_{\pm.002}$ \\
\bottomrule
\end{tabular}
\caption{Overall micro results, averaged over 3 runs. Frames reports accuracy. Extraction and relations report P/R/$F_1$. \textendash{}: not applicable.}
\label{tab:overall}
\end{table}

\paragraph{Per-country generalization.}
Table~\ref{tab:percountry} shows the country-level breakdown. Frame accuracy is the most stable metric across jurisdictions, ranging from $0.82$ (Australia) to $0.90$ (USA), an $8$-point spread that confirms the binary deliberative--managerial frame separation generalizes beyond the FEMA-centric setting in which it was developed. Relation $F_1$ shows a wider but coherent pattern. USA leads at every nesting level (Detection $0.77$, Subtype $0.65$), while UK trails (Detection $0.70$, Subtype $0.53$). We attribute the USA lead to the fact that the rule set was inductively refined on FEMA documents, which are dense, self-referential, and rich in the explicit modality and agency markers our features consume. UK text, by contrast, is shorter and more horizontally-linked across the Civil Contingencies Secretariat, Local Resilience Forums, and devolved-nation agencies. This distributed governance structure produces fewer and shorter argument chains ($2.8$ rel/arg, vs.\ $7.9$ in the USA), and it fragments the agency typology across multiple actor types. Crucially, the cross-country gap on relations ($0.07$ at Detection, $0.12$ at Subtype) is small enough that we read it as graceful degradation under genuine corpus shift, not as failure to generalize.

\begin{table}[t]
\centering
\begin{tabular}{@{}l|cc|ccc@{}}
\toprule
\textbf{Country} & \textbf{Ext} & \textbf{Frm} & \textbf{Det} & \textbf{Pol} & \textbf{Sub} \\
\midrule
Australia & \textbf{.95} & .82 & .72 & .64 & .54 \\
Canada    & .91 & .88 & .71 & .64 & .56 \\
UK        & .82 & .84 & .70 & .61 & .53 \\
USA       & .92 & \textbf{.90} & \textbf{.77} & \textbf{.70} & \textbf{.65} \\
\midrule
\textbf{Micro} & \textbf{.91} & \textbf{.86} & \textbf{.73} & \textbf{.66} & \textbf{.58} \\
\bottomrule
\end{tabular}
\caption{Per-country breakdown, $n$-weighted across sub-documents. \emph{Ext}, \emph{Det}, \emph{Pol}, and \emph{Sub} report $F_1$ (extraction and the Detection, Polarity, and Subtype relation levels). \emph{Frm} reports accuracy.}
\label{tab:percountry}
\end{table}

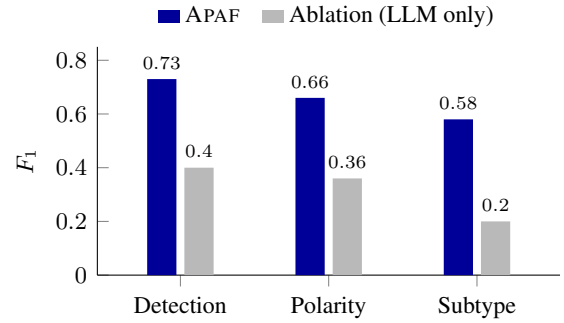
\begin{figure}[t]
\centering
\begin{tikzpicture}
  \begin{axis}[
    ybar=3pt,
    width=\linewidth,
    height=4.6cm,
    bar width=11pt,
    ymin=0, ymax=0.85,
    ytick={0,0.2,0.4,0.6,0.8},
    ylabel={$F_1$},
    ylabel style={font=\small},
    symbolic x coords={Detection, Polarity, Subtype},
    xtick=data,
    xticklabel style={font=\small},
    yticklabel style={font=\footnotesize},
    nodes near coords,
    nodes near coords style={font=\scriptsize},
    enlarge x limits=0.28,
    legend style={
      at={(0.5,1.04)}, anchor=south,
      legend columns=2,
      font=\small,
      draw=none,
      /tikz/every even column/.append style={column sep=0.6em},
    },
    legend image code/.code={%
      \draw[#1, draw=none] (0cm,-0.08cm) rectangle (0.28cm,0.08cm);%
    },
    axis y line*=left, axis x line*=bottom,
  ]
  \addplot[fill=blue!60!black, draw=none] coordinates {
    (Detection,0.73) (Polarity,0.66) (Subtype,0.58)
  };
  \addplot[fill=gray!55, draw=none] coordinates {
    (Detection,0.40) (Polarity,0.36) (Subtype,0.20)
  };
  \legend{\textsc{Apaf}, Ablation (LLM only)}
  \end{axis}
\end{tikzpicture}
\caption{Relation $F_1$ at each nesting level: \textsc{Apaf} versus the LLM-only ablation. The 30 to 38 $F_1$-point gap widens with granularity.}
\label{fig:ablation-chart}
\end{figure}

\paragraph{Ablation.}
We test whether the deterministic rules in Step~4 contribute signal beyond what a prompted LLM could produce. The ablation replaces Step~4 with per-source LLM calls. For each source argument, the LLM receives the source text, all other argument texts in the sub-document as candidate targets, and the document's topic argument as context. For each target, it returns a relation type from \{none, support, attack\} and a subtype from the four classes the rules produce. The rest of the pipeline is unchanged. Annotated arguments are fed in as in the relation evaluation, and Steps 2--3 compute frames and features in exactly the same way. The LLM sees only argument texts, never the frames or features. This setup directly tests whether a prompted LLM, given the subtype labels and short definitions, can replace the symbolic structure on this task. Figure~\ref{fig:ablation-chart} reports the comparison on the full 100 sub-document dataset.

The rules outperform the LLM ablation at every level, by 30 to 38 $F_1$ points. The gap widens with granularity. On Detection the LLM reaches 55\% of the rule-based $F_1$, on Polarity 55\%, and on Subtype only 34\%. We attribute this to two factors. First, the LLM is precision-leaning and recall-low. It predicts roughly 5{,}400 relations against the rules' 7{,}700, and identifies about 35\% of annotated pairs against the rules' 75\%. Second, the LLM degrades sharply at the Subtype level across all four countries (Subtype $F_1$ in the $0.18$ to $0.22$ range), while the rules retain country-specific signal (Subtype $F_1$ from $0.53$ in UK to $0.65$ in USA). This is consistent with the design claim of the framework. The rules encode the frame-mediated semantics that classical argument mining collapses, and a prompted LLM, asked to choose among the four subtype labels, does not recover comparable structure even when given short subtype definitions in the prompt. We report an error analysis in Appendix~\ref{app:error-analysis}.

\section{Discussion}
\label{sec:discussion}

\textsc{Apaf} is, to our knowledge, the first system evaluated on a typed frame-mediated relation classification task over real disaster-governance policy. The four frame-mediated subtypes (agency reduction, agenda shift, instrumental support, normative support) and the four-country governance corpus together define a task that did not previously have a corpus-level benchmark. The reported numbers (0.73 Detection $F_1$, 0.58 at the Subtype level, 0.86 frame accuracy) therefore establish a baseline rather than a peak. The 30 to 38 $F_1$-point gap over the LLM-only ablation shows that the deterministic frame-mediated rules are doing real work, and the small cross-country gaps (0.07 at Detection, 0.12 at Subtype) show that the rule design generalizes beyond its FEMA-centric origins. We expect future systems to close the Polarity-to-Subtype drop, target the residual UK gap, or extend the corpus to non-anglophone jurisdictions.

Critical discourse analysis of policy text has long argued that governance discourse contains tensions between a communicative rationality oriented toward participation and public deliberation, and an instrumental rationality oriented toward administration, implementation efficiency, and compliance \citep{habermas1984theory,fairclough2003analysing,arnstein1969ladder}. The claim has rested on close readings of small selections of documents. Our corpus statistics (\S\ref{sec:dataset}) and relation-level results (\S\ref{sec:experiments}) jointly support the observation at scale. Instrumental support relations dominate normative support across all four jurisdictions. Agenda-shift attacks recur with similar density in every national setting. Frame distributions are stable across countries which share similar policy discourses, despite some variety in institutional structures. The system thus provides quantitative grounding for a body of claims that has typically been resistant to operationalization.

In the context of public goods, disaster governance functions as a classical collective-action domain where outcomes depend on shared trust, coordinated participation, and legitimacy \citep{undrr2015sendai}. By transforming opaque policy texts into structured, frame-mediated argumentation graphs, \textsc{Apaf} increases the transparency and contestability of institutional reasoning, enabling a wide range of actors to inspect how agency and responsibility are allocated across competing rationalities. In this sense, explainable discourse representation directly supports democratic oversight by making the reasoning behind public risk governance auditable, comparable, and open to challenge.
We do not claim deliberative framing is uniformly preferable to managerial framing, only that the gap between them is structural, measurable, and amenable to NLP analysis. Civic-tech, deliberative-platform \citep{delgado2023participatory}, and policy-tracking \citep{zolkowski2022climate,sietsma2024machine} projects can build on this infrastructure without inheriting our normative position, by re-typing relations or substituting frame typologies suited to their domain.

\section{Conclusion}
\label{sec:conclusion}

We presented \textsc{Apaf}, a hybrid LLM--symbolic pipeline that constructs Quantitative Bipolar Argumentation Frameworks from disaster-governance policy text, introducing a typology of frame-mediated attack and support relations that classical argument mining does not model. On a new dataset of 100 sub-documents spanning the USA, UK, Canada, and Australia, the pipeline reaches $0.91$ extraction $F_1$, $0.86$ frame accuracy, and $0.73$ Detection $F_1$ ($0.58$ at the Subtype level), with stable performance across jurisdictions. The output is interpretable by construction: every relation and weight is traceable to a rule or recorded LLM justification, and the human-in-the-loop interface lets domain experts modify and re-solve the graph in a single pass. Future work includes extension to non-anglophone governance corpora, integration with deliberative platforms as an analytical layer, and richer frame typologies for governance domains beyond disaster risk reduction.

\section*{Limitations}

Our work has several limitations that constrain the scope of its claims. First, the corpus is restricted to anglophone democracies with broadly similar institutional structures and a shared Sendai engagement. The frame typology, the agency categories, and the discourse lexica were inductively refined on this slice, and generalization to non-anglophone, non-Westminster, or non-federal contexts is an open empirical question. Second, the annotation was performed by two researchers with reconciliation by discussion rather than by a larger independent panel with a reported inter-annotator agreement scalar. We made this choice deliberately because a single IAA number for a four-layer structured task (spans, frames, relations, topic) would obscure more than it reveals, but it does limit external statistical validation. Third, our pipeline relies on an LLM for argument extraction, agency typing, and contradiction counting. While the symbolic layers above are deterministic and inspectable, LLM-driven errors propagate downstream, and we have not measured their marginal contribution to the Polarity and Subtype relation error in isolation. Fourth, the ``explainability'' \textsc{Apaf} delivers is provenance-traceability for every output, not algorithmic interpretability of the LLM's internal computations. Users can audit why a rule fired but not why the LLM produced a particular feature value. Fifth, the binary frame typology (deliberative or managerial) is by design a strong simplification. Governance discourse contains additional rationalities, such as technocratic, market-liberal, and civic-republican framings, that a richer typology would surface but that we have not attempted to model.

\section*{Ethics Statement}

The corpus consists of publicly available policy documents released by national and sub-national governments, and no personal data is involved. Our framework is interpretive rather than normative. The codebook reflects an analytical perspective grounded in critical discourse analysis, and the system's outputs should be read as evidence to inform human deliberation rather than as objective measurements of policy quality. We see two principal dual-use risks. First, a system that surfaces implicit framing dynamics can in principle be used in reverse, to rewrite policy text so that it scores well on this or a related framework while preserving substantively top-down structures. We mitigate this only weakly by releasing the codebook openly so that such gaming is itself contestable. Second, automated framing analysis applied to small samples or used as a sole input to procurement, funding, or audit decisions could entrench rather than disrupt power asymmetries. We therefore emphasize the human-in-the-loop interface as a required component, not an optional one, and we frame the system as analytical infrastructure for participatory governance rather than as autonomous decision support. We follow recent calls for explicit harm taxonomies for algorithmic systems \citep{shelby2023sociotechnical,weidinger2022taxonomy} and for civic-participation-aware AI design \citep{delgado2023participatory,robertson2020what} in shaping both these mitigations and our broader research agenda.

\bibliography{custom}

\clearpage 

\appendix

\section{Interactive Interface and Worked Example}
\label{app:interface}

\begin{figure}[!ht]
\centering
\includegraphics[width=\linewidth]{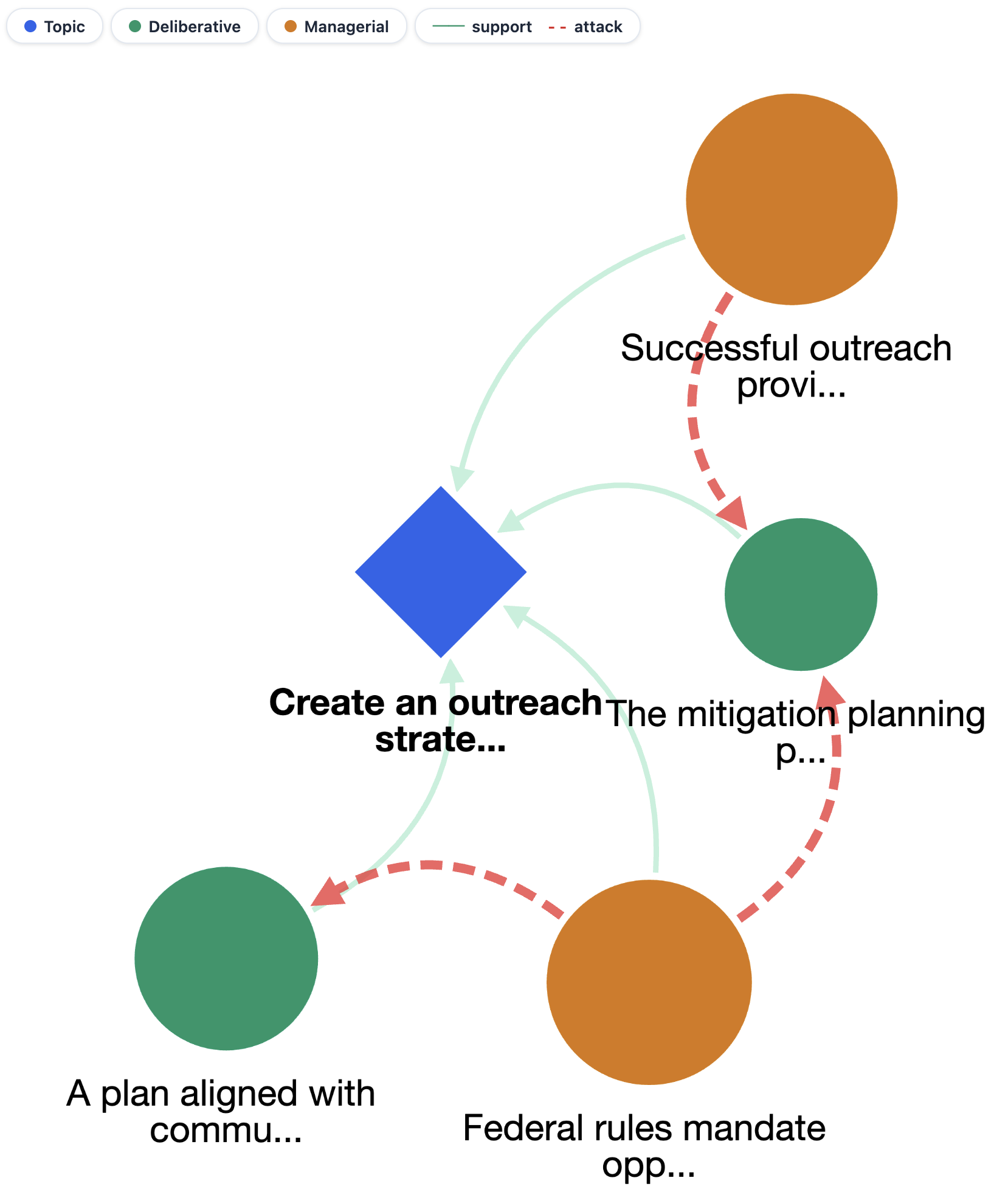}
\caption{A five-argument subset of the FEMA Task~3 output, chosen to illustrate the four frame-mediated relation subtypes. The topic argument (blue diamond) receives normative support from two deliberative arguments (green) and instrumental support from two managerial arguments (orange), which also attack the deliberative arguments through \emph{agency reduction} and \emph{agenda shift} (dashed red edges).}
\label{fig:interface}
\end{figure}

\begin{figure*}[t]
\centering
\includegraphics[width=\linewidth]{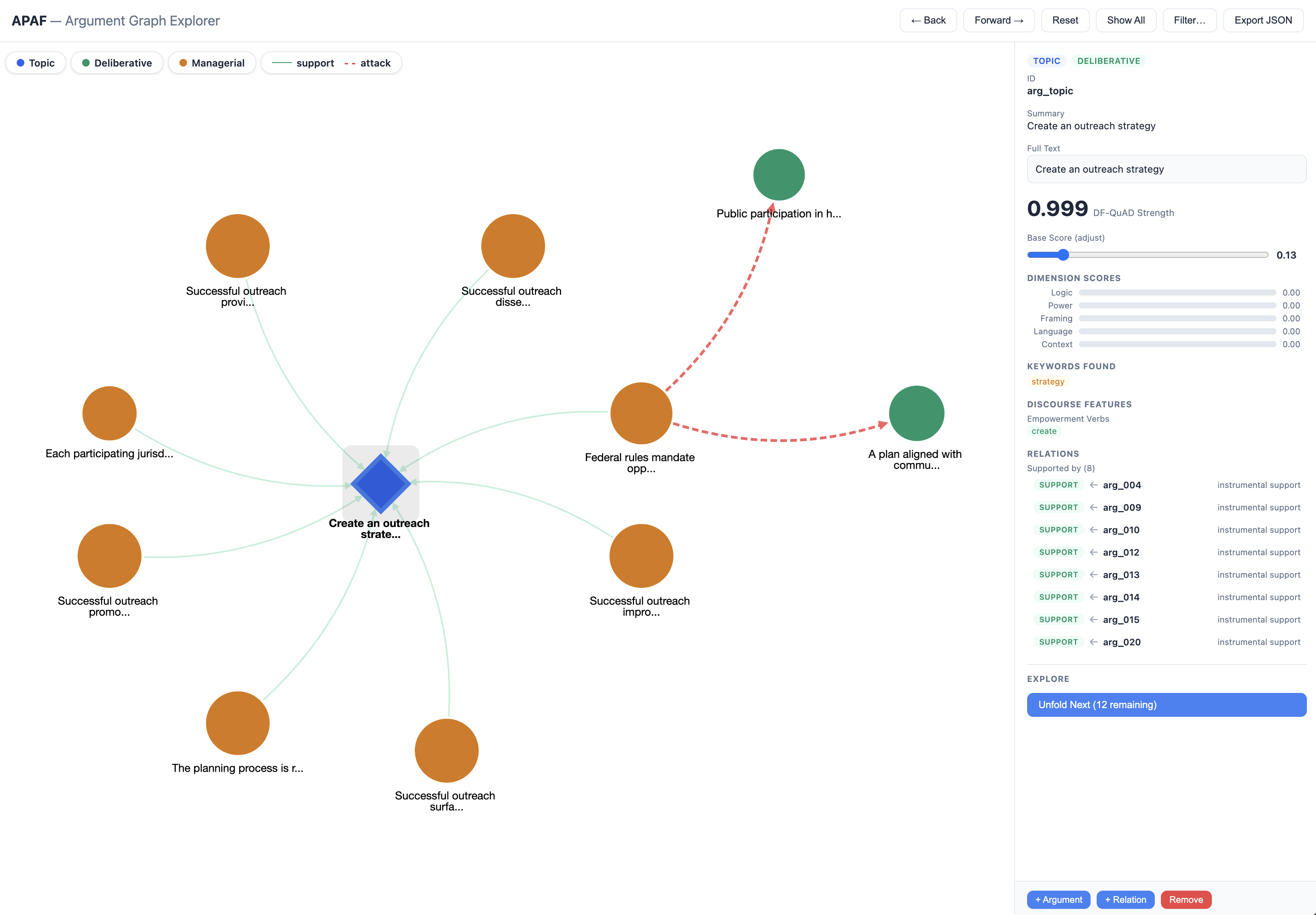}
\caption{The full \textsc{Apaf} interactive graph explorer. The graph view (left and center) renders arguments as nodes coloured by frame (blue diamond for the topic, green for deliberative, orange for managerial), with solid teal edges for supports and dashed red edges for attacks. The right panel shows details of the selected node (here the topic argument), including the DF-QuAD strength $\sigma$, the base score $\tau$ with sliders for the five weight dimensions, the keywords and features that drove classification, and the list of incoming and outgoing relations with their subtypes. The toolbar at the top supports navigation, filtering, and JSON export.}
\label{fig:full-interface}
\end{figure*}

\textsc{Apaf}'s output is displayed in an interactive graph explorer that a domain expert uses to inspect, edit, and re-solve the QBAF in a single pass. Figure~\ref{fig:full-interface} shows the full layout on the FEMA Task~3 outreach-strategy document.

\paragraph{Layout.}
The graph view renders arguments as nodes coloured by frame and relations as typed edges. The right panel shows details for the selected node, including the argument text, the frame label, the current DF-QuAD strength $\sigma(a)$, the base score $\tau(a)$ with a slider for direct editing, and per-dimension sub-scores (logic, power, framing, language, context). The right panel also lists the detected keywords and discourse features that the symbolic rules consumed, and enumerates the relations the node participates in together with their subtypes. The toolbar supports navigation (Back, Forward, Reset), filtering to a user-specified subset of arguments, JSON export, and the Show All and Unfold-from-node operations that let an expert progressively explore a large argumentation graph.

\paragraph{Editing.}
Four edit operations are accessed from the right panel and the bottom-right action buttons. Promoting an implicit argument triggers re-embedding and rule re-firing without re-extracting from text. Flipping the frame of an argument reuses cached LLM contradiction counts and embeddings. Adjusting the base score or any dimension score updates $\tau(a)$ in place. Adding or removing an argument or relation modifies the graph. After every change the strengths $\sigma$ are re-solved deterministically by DF-QuAD, so the consequences of each edit are immediately visible.

\paragraph{Worked example.}
Figure~\ref{fig:interface} shows a five-argument subset of the same FEMA Task~3 output, chosen to illustrate the four frame-mediated relation subtypes in a compact layout. The topic argument (blue diamond, ``Create an outreach strategy'') receives normative support from two deliberative arguments (green) and instrumental support from two managerial arguments (orange). The same managerial arguments simultaneously attack the deliberative ones through \emph{agency reduction} and \emph{agenda shift} (dashed red edges), making the central participatory--managerial tension visible at a glance. Selecting any edge in the interface reveals the rule that produced it and the LLM call or lexicon match that supplied each feature.

\section{LLM Prompts}
\label{app:prompts}

\textsc{Apaf} invokes an LLM at four points in the pipeline: argument extraction (Step~1), frame tie-breaking on ambiguous arguments (Step~2), agency typing on arguments where the keyword scan is inconclusive (Step~3), and internal-contradiction counting for the \emph{logic} dimension of the base score (Step~5). The system prompts for these four calls are reproduced below. Each prompt is paired with a structured-output schema; we use OpenAI's structured-output interface so the model returns parsed JSON that matches the schema. We omit the user prompts, which are short templates that splice the argument text into a JSON envelope.

\subsection{Argument extraction (Step~1)}
\label{app:prompt-extraction}

{\footnotesize
\begin{verbatim}
You are an expert policy analyst. Your
task is to identify ARGUMENTS that are
EXPLICITLY stated in a policy document.

DEFINITION. An ARGUMENT is a complete,
stance-bearing proposition explicitly
stated in the text. To qualify, a span
must satisfy all four criteria:
 (a) PROPOSITION. The span asserts
     something is, should be, will be,
     or must be the case. It has a
     subject and a predicate.
 (b) STANCE. The span makes a claim a
     reader could agree with, disagree
     with, support, or contest.
 (c) EXPLICIT. The claim is present in
     the text. Do not infer from
     background knowledge.
 (d) STAND-ALONE. The claim still makes
     sense if extracted and shown to a
     reader unfamiliar with the
     surrounding paragraph.

ARGUMENT TYPES. Choose exactly one per
argument:
 - SUBSTANTIVE: value claims, causal
   claims, justifications, claims about
   consequences, stakes, or beneficiaries.
 - PROCEDURAL: process specifications,
   methods, institutional arrangements,
   mechanisms.

NON-ARGUMENTS. Skip: section headings,
cross-references, isolated definitions,
pure descriptive facts, sentence
fragments, background narrative, and
citations.

GRANULARITY. One sentence with one claim
= one argument. Bulleted lists under an
inheriting heading produce one argument
per bullet, with the inherited subject
combined with each bullet's predicate.
Do not split conditionals.

LANGUAGE PRESERVATION. Preserve the
claim's wording. Do not change verbs,
modal verbs (must, may, should, shall,
will), or noun phrases.

TOPIC EXCLUSION. The topic argument is
provided separately as context. Do not
extract it.

OUTPUT. For each argument, emit arg_id,
text, summary, arg_type, source_quote.
\end{verbatim}
}

\subsection{Frame tie-breaker (Step~2)}
\label{app:prompt-frame}

Called only on arguments where the deliberative and managerial keyword scans return tied or near-tied scores. Roughly $15$--$25\%$ of arguments per sub-document trigger this call.

{\footnotesize
\begin{verbatim}
You are classifying short policy-text
arguments into one of two governance
frames.

The text comes from a hazard-mitigation
policy document in one of four countries
(USA, UK, Canada, Australia). The
underlying frame distinction is the SAME
across countries; recognize country-
specific equivalents as instances of the
signals below.

DELIBERATIVE: process-oriented, values-
driven, community-agency. Signals:
 - Community / public participation,
   engagement, inclusion, consultation
 - Local knowledge, lived experience
 - Residents, whole-community, locals
   as the actor
 - Transparency, equity, collaborative
 - Empowerment, capacity-building,
   community-led decisions
 - Two-way / horizontal / shared agency

MANAGERIAL: compliance, authority,
implementation. Signals:
 - Regulatory or jurisdictional
   requirements, statutory law
 - Top-down authority as the acting
   agent (FEMA, Council, state agency)
 - Technical procedures: risk assessment,
   mitigation actions, data analysis
 - Formal hierarchies, directed agency
 - Output-oriented success metrics

KEY DISTINCTION. Authority MENTION alone
is not enough to make an argument
managerial. The frame is determined by
WHO ACTS and WHAT KIND of agency is at
stake, not by which names appear. If an
authority is mentioned but the substance
is about community input, classify as
DELIBERATIVE.

For each argument, decide the SINGLE
frame that best characterizes it. If
genuinely mixed, choose the dominant one.
\end{verbatim}
}

\subsection{Agency typing (Step~3)}
\label{app:prompt-agency}

Called only on arguments where the keyword scan over the six agency categories returns no clear winner. The prompt receives both the argument text and the keyword matches found for each category.

{\footnotesize
\begin{verbatim}
You classify the AGENCY TYPE of policy
arguments. Agency = who is presupposed
to act in the argument.

Choose ONE category per argument:

unilateral_top: top-down national /
  federal authority acting unilaterally.
  USA: FEMA, federal agencies.
  UK:  central government, Defra, the
       Cabinet Office.
  CAN: Public Safety Canada.
  AUS: the Commonwealth.

unilateral_mid: mid-level unilateral
  authority (state / regional / juris-
  dictional), or 'you' acting alone.
  USA: the State, jurisdictional
       authorities.
  UK:  Environment Agency, LLFA,
       the Council.
  CAN: the City, provincial government.
  AUS: state agency, planning authority.

directed: a specific directed actor
  carrying out instructions (planning
  teams, departments, local leaders).

shared_symmetrical: shared / symmetric
  agency among community, neighbours,
  the public, residents, stakeholders.

horizontal: mutual, reciprocal, two-way,
  among equals.

none: no clear agent.

Keyword matches are EVIDENCE, not a
verdict. You may override based on
semantics. Authority mentions that act
as references (not as the acting
subject) do not select unilateral_top.
\end{verbatim}
}

\subsection{Contradiction counting (Step~5)}
\label{app:prompt-contradiction}

Used to score the \emph{logic} dimension of the base weight $\tau(a)$.

{\footnotesize
\begin{verbatim}
You are a logic analyst. For each
argument, count the number of internal
contradictions: places where the
argument's own premises conflict with
its conclusion, or different parts of
the argument assert incompatible things.

Count only genuine logical
contradictions, not merely tensions or
qualifications.

For each argument, return arg_id,
contradiction_count (integer >= 0), and
a brief justification.
\end{verbatim}
}

\section{Keyword Lexicons}
\label{app:keywords}

This appendix lists the keyword lexicons that drive Steps~2--5 of the pipeline. The lists were developed inductively from the FEMA Local Mitigation Planning Handbook and the Sendai Framework, then expanded with UK / Canada / Australia equivalents during annotation reconciliation. Matching is case-insensitive and uses word-boundary checks; multi-word keys match across whitespace.

\subsection{Frame keywords (Step~2)}
\label{app:kw-frames}

\paragraph{Deliberative.}
Three sub-categories. Below we show a representative subset; the full lists contain $185$ deliberative keywords across the three categories.

\noindent\emph{Process orientation:}
{\small\itshape
public participation, public engagement, public comment, community engagement, community-led, community-driven, community develops, communities decide, co-create, collaborative, working together, bring people together, shared understanding, engagement process, joint planning, ongoing dialogue, mutual aid, peer-to-peer, broad participation, communicate openly, foster relationship, empowering communities, empowering individuals, proactive planning, build community resilience, increase local resilience, drive decisions, support strengthening, share threat, share views, share ideas, intent to participate, communities take actions, asynchronous hybrid meeting, in-person public meetings, public workshops, public review.}

\noindent\emph{Values:}
{\small\itshape
community, resilient, transparency, lived experience, community-driven, community-based, local knowledge, local context, local culture, community values, community priorities, community vision, community well-being, civic engagement, inclusive, meaningful engagement, equal access, equal opportunities, shared understanding, empowered communities, mutual, longstanding community values, bottom-up approach, evidence-based knowledge.}

\noindent\emph{Agency (deliberative actors):}
{\small\itshape
whole-community, public, everyone, neighbours, locals, residents, individuals, small communities, community-based organizations, community members, all stakeholders, wide range of stakeholders, each community, grassroots groups, local community groups, all interested parties, communities of practice, hazard-specific coalitions, individuals, families, and households.}

\paragraph{Managerial.}
Four sub-categories with $190$ total keywords.

\noindent\emph{Compliance:}
{\small\itshape
regulation, require, federal requirements, must, federal regulation, federal law, ordinances, FEMA approval, FEMA regulation, federal agency, eligible, eligible mitigation plan, awarded funds, grant program, mitigation-specific, meet the requirements, take the lead, to approve, federally recognized, ready-made opportunity, jurisdictional boundaries.}

\noindent\emph{Authority:}
{\small\itshape
jurisdiction, federal, FEMA, the State, federal government, federal agencies, local government, decision makers, multi-jurisdictional, government entity, agency, plan owner, governing body, professional planning team, coordinating structures, national, state mitigation planning, council, committee, comission.}

\noindent\emph{Implementation:}
{\small\itshape
emergency management, mitigation actions, management, funding, securing funding, carry out, data analysis, hiring a contractor, procurement, risk assessment, technical information, planning mechanisms, formal, blueprint, project development, to leverage, to organize, to evaluate, to assess, to administer, to track, to coordinate, to define, to identify, to enforce, enforcement, the precondition, planning area, maintenance cycle.}

\noindent\emph{Success metrics:}
{\small\itshape
buy-in, generate support, strengthen outcomes, maximizes agreement, greater acceptance, agreement, works best.}

\subsection{Agency lexicons (Step~3)}
\label{app:kw-agency}

Six agency types, each with a small dedicated lexicon. Detection is regex-based with an LLM fallback (Appendix~\ref{app:prompt-agency}) when the regex is inconclusive.

\begin{itemize}\setlength\itemsep{0.15em}
\item \textbf{unilateral\_top:} {\small\itshape FEMA, Federal, national, federal government, federal agencies.}
\item \textbf{unilateral\_mid:} {\small\itshape you, jurisdiction, state authority, state mitigation planning, local government, decision makers, multi-jurisdictional, single jurisdiction, government entity, agency, authorities, governing body, the state, coordinating structures, community officials, committee, council, comission, organisation, response services.}
\item \textbf{directed:} {\small\itshape the planning team, departments, local leaders, committee, local partners, leaders, emergency managers, plan owner, customers, experts, consultants, professional planning team, support services, team.}
\item \textbf{shared\_symmetrical:} {\small\itshape whole-community, public, locals, individuals, residents, neighbors, everyone, social networks, neighboring communities, community-based organizations, community members, all stakeholders, wide range of stakeholders, grassroots groups, all community members, individuals families and households, community, communities.}
\item \textbf{horizontal:} {\small\itshape mutual, reciprocal, two-way, bidirectional, among equals.}
\item \textbf{none:} no clear agent matches.
\end{itemize}

\subsection{Verb classes (Step~3)}
\label{app:kw-verbs}

Verb classes signal the relational stance of the argument: tokenistic verbs reduce participation to passive consultation, while empowerment verbs assign genuine agency to the community.

\begin{itemize}\setlength\itemsep{0.15em}
\item \textbf{Tokenistic verbs} ({\small\itshape consume into rule A1}): to inform, to educate, invite parties, share data, sending out, holding public meetings, to explain, to determine, to include, to update, to coordinate, to administer, to help, to organize, to form, to facilitate, to oversee, to regulate, to teach, to give chances, to hold the meeting, to set up priorities, to address, must be informed, be informed, to enforce, to integrate.
\item \textbf{Empowerment verbs} ({\small\itshape consume into rule A1}): community accomplishes, have a voice, enable, community undertakes actions, community builds resilience, community takes action, encourage, create, community develops, collaborate, to discuss, communities may choose, community decides, communities should review, to participate, to motivate, share threat, share views, share their views and ideas, the public to review and comment, share ideas, accommodate the needs, to contribute, defined by the community, community-led, communities choose, building capacity, drive decisions, community will accomplish, communities consider, evaluate the alternatives, advance resilience, working together, communities promote, communities implement, communicate openly, foster relationship.
\end{itemize}

\subsection{Instrumental markers (Step~3)}
\label{app:kw-instrumental}

Two lists corresponding to the two directionalities of the \emph{instrumental support} rule (S1 and S3). A managerial argument carrying an item from the first list supports a deliberative target instrumentally; a managerial-to-managerial pair where the source carries an item from the second list and the two are close in embedding space fires the same subtype with a managerial target.

\begin{itemize}\setlength\itemsep{0.15em}
\item \textbf{instrumental\_for\_deliberative}: {\small\itshape to require, to meet certain requirements, strengthen outcomes, generate support, agreement, acceptance, achieve, blueprint, federal regulations, finding funding, pursuing funding, to secure funding, funding, buy-in, quality of the plan, maximizes chances, greater acceptance, reducing losses, must, to reduce, reduce workloads.}
\item \textbf{instrumental\_for\_managerial}: {\small\itshape carry out, ensure, employ, identify, adopting, assess risk, eliminate risk, to apply, to eliminate, to complete, to conduct, lessen the impact, to assess, to focus, to speed up, to evaluate, receive grant, to track, to leverage, expertise, share expertise, you build, to describe, to define, to bring the plan, to agree, to take advantage, to establish, to promote, works best.}
\end{itemize}

\subsection{Modality (Step~5, language dimension)}
\label{app:kw-modality}

Used to score the \emph{language} dimension of $\tau(a)$.

\begin{itemize}\setlength\itemsep{0.15em}
\item \textbf{Strong modality} ($\to 1.0$): must, shall, require, will, mandate.
\item \textbf{Weak modality} ($\to 0.5$): may, could, might.
\item \textbf{Absent} ($\to 0.3$): none of the above present.
\end{itemize}

\section{Deterministic Rules}
\label{app:rules}

This appendix lists the seven deterministic rules that produce relations in Step~4, together with the scoring functions for the five base-score dimensions in Step~5. All rules fire over directed argument pairs $(a_s, a_t)$. At most one relation is produced per pair. Within a pair, attacks are checked before supports, and rules fire in the order $A1 \to A2 \to S1 \to S2 \to S3$. The topic argument $a_0$ is never a source and is never the target of an attack; topic targets are handled by the two dedicated topic rules (TOPIC\_N, TOPIC\_S) which replace all other rules for those pairs.

\subsection{Notation}
\label{app:rules-notation}

For an argument $a$, $\phi(a) \in \{\mathrm{D}, \mathrm{M}\}$ is its frame (deliberative or managerial). $\alpha(a)$ is its agency type, drawn from the six categories in Appendix~\ref{app:kw-agency}. We use the shorthand $\alpha_M = \{\text{unilateral\_top}, \text{unilateral\_mid}, \text{directed}\}$ and $\alpha_D = \{\text{shared\_symmetrical}, \text{horizontal}\}$ to group the managerial-side and deliberative-side agency types. $\nu(a)$ is the set of verb-class signals carried by $a$, a subset of $\{\text{tokenistic}, \text{empowerment}\}$. $\iota(a)$ is the set of instrumental-marker signals, a subset of $\{\text{for-D}, \text{for-M}\}$. $\mathbf{e}(a) \in \mathbb{R}^{384}$ is the sentence embedding, and $\theta = 0.6$ is the cosine threshold.

\subsection{Attack rules}
\label{app:rules-attacks}

\paragraph{A1 --- Agency reduction.}
\begin{quote}\small
\textbf{IF}~~$\phi(a_s) = \mathrm{M}$\\
\textbf{AND}~tokenistic $\in \nu(a_s)$\\
\textbf{AND}~$\phi(a_t) = \mathrm{D}$\\
\textbf{AND}~empowerment $\in \nu(a_t)$\\
\textbf{THEN}~emit attack$(a_s, a_t)$\\
\hphantom{THEN~}of subtype agency-reduction.
\end{quote}

\paragraph{A2 --- Agenda shift.}
\begin{quote}\small
\textbf{IF}~~$\phi(a_s) = \mathrm{M}$\\
\textbf{AND}~$\alpha(a_s) \in \alpha_M$\\
\textbf{AND}~$\phi(a_t) = \mathrm{D}$\\
\textbf{AND}~$\alpha(a_t) \in \alpha_D$\\
\textbf{THEN}~emit attack$(a_s, a_t)$\\
\hphantom{THEN~}of subtype agenda-shift.
\end{quote}

\subsection{Support rules}
\label{app:rules-supports}

\paragraph{S1 --- Instrumental support, deliberative target.}
\begin{quote}\small
\textbf{IF}~~$\phi(a_s) = \mathrm{M}$\\
\textbf{AND}~for-D $\in \iota(a_s)$\\
\textbf{AND}~$\phi(a_t) = \mathrm{D}$\\
\textbf{THEN}~emit support$(a_s, a_t)$\\
\hphantom{THEN~}of subtype instrumental.
\end{quote}
\noindent No vector check (rule priority).

\paragraph{S2 --- Normative support.}
\begin{quote}\small
\textbf{IF}~~$\phi(a_s) = \mathrm{D}$\\
\textbf{AND}~$\phi(a_t) = \mathrm{D}$\\
\textbf{AND}~$\cos(\mathbf{e}(a_s), \mathbf{e}(a_t)) \geq \theta$\\
\textbf{THEN}~emit support$(a_s, a_t)$\\
\hphantom{THEN~}of subtype normative.
\end{quote}

\paragraph{S3 --- Instrumental support, managerial target.}
\begin{quote}\small
\textbf{IF}~~$\phi(a_s) = \mathrm{M}$\\
\textbf{AND}~for-M $\in \iota(a_s)$\\
\textbf{AND}~$\phi(a_t) = \mathrm{M}$\\
\textbf{AND}~$\cos(\mathbf{e}(a_s), \mathbf{e}(a_t)) \geq \theta$\\
\textbf{THEN}~emit support$(a_s, a_t)$\\
\hphantom{THEN~}of subtype instrumental.
\end{quote}

\subsection{Topic rules}
\label{app:rules-topic}

The two topic rules fire whenever the target is the topic argument $a_0$. They replace all other rules for that pair and require no feature or vector check.

\paragraph{TOPIC-N --- Normative support toward topic.}
\begin{quote}\small
\textbf{IF}~~$a_t = a_0$\\
\textbf{AND}~$\phi(a_s) = \mathrm{D}$\\
\textbf{THEN}~emit support$(a_s, a_0)$\\
\hphantom{THEN~}of subtype normative.
\end{quote}

\paragraph{TOPIC-S --- Instrumental support toward topic.}
\begin{quote}\small
\textbf{IF}~~$a_t = a_0$\\
\textbf{AND}~$\phi(a_s) = \mathrm{M}$\\
\textbf{THEN}~emit support$(a_s, a_0)$\\
\hphantom{THEN~}of subtype instrumental.
\end{quote}

\subsection{Priority and tie-breaking}
\label{app:rules-priority}

Within a single ordered pair $(a_s, a_t)$, the rule engine checks topic, then attacks ($A1, A2$), then supports ($S1, S2, S3$) in that order, stopping at the first match. When two attack rules both match the same pair (typical when a managerial argument with tokenistic verbs targets a deliberative argument with shared agency), $A1$ takes priority. This priority is conservative: $A1$ is the stronger semantic claim (active reduction of participation) and $A2$ is the weaker structural claim (agency reassignment).

\subsection{Base-score functions (Step~5)}
\label{app:rules-weights}

The base score $\tau(a) = \tfrac{1}{5}\sum_{d \in \mathcal{D}} w_d(a)$ averages five dimensions. Each dimension function is either a deterministic lookup or a deterministic count over keyword matches.

\paragraph{Logic.} A decreasing function of the LLM-counted contradictions $c(a)$:
\begin{itemize}\setlength\itemsep{0pt}
\item $c(a) = 0$ $\to$ $w_{\text{logic}} = 1.0$
\item $c(a) = 1$ $\to$ $w_{\text{logic}} = 0.7$
\item $c(a) \geq 2$ $\to$ $w_{\text{logic}} = 0.3$
\end{itemize}

\paragraph{Power.} A frame-conditioned lookup over the agency type $\alpha(a)$. For managerial arguments:
\begin{itemize}\setlength\itemsep{0pt}
\item $\alpha = $ unilateral\_top $\to$ $w_{\text{power}} = 1.0$
\item $\alpha = $ unilateral\_mid $\to$ $w_{\text{power}} = 0.7$
\item $\alpha = $ directed $\to$ $w_{\text{power}} = 0.4$
\item otherwise $\to$ $w_{\text{power}} = 0.2$
\end{itemize}
For deliberative arguments:
\begin{itemize}\setlength\itemsep{0pt}
\item $\alpha = $ shared\_symmetrical $\to$ $w_{\text{power}} = 0.7$
\item $\alpha = $ horizontal $\to$ $w_{\text{power}} = 0.4$
\item otherwise $\to$ $w_{\text{power}} = 0.2$
\end{itemize}

\paragraph{Framing.} The density of frame-consistent keywords in $a$. Let $k(a)$ be the count of keyword matches from the lexicon for $\phi(a)$. Then $w_{\text{framing}}(a) = \min(1.0,\, k(a)/5)$.

\paragraph{Language.} The strongest modal verb in $a$ (modality lexicons in Appendix~\ref{app:kw-modality}):
\begin{itemize}\setlength\itemsep{0pt}
\item strong modality present $\to$ $w_{\text{lang}} = 1.0$
\item weak modality present $\to$ $w_{\text{lang}} = 0.5$
\item no modality $\to$ $w_{\text{lang}} = 0.3$
\end{itemize}

\paragraph{Context.} For managerial arguments, $m(a)$ counts the managerial-keyword matches; for deliberative arguments, $m(a)$ counts the Sendai/DRR-lexicon matches.
\begin{itemize}\setlength\itemsep{0pt}
\item $m(a) \geq 3$ $\to$ $w_{\text{ctx}} = 1.0$
\item $1 \leq m(a) \leq 2$ $\to$ $w_{\text{ctx}} = 0.6$
\item $m(a) = 0$ $\to$ $w_{\text{ctx}} = 0.2$
\end{itemize}

\paragraph{Topic base score.} The topic argument $a_0$ receives a fixed base score $\tau(a_0) = 0.5$, treated as a hyperparameter.

\section{Extended Error Analysis}
\label{app:error-analysis}

We expand the brief error analysis of \S\ref{sec:experiments} with quantitative breakdowns and representative examples for each of the five recurring failure modes.

\paragraph{E.1 Sparse-relation sub-documents.}
Nine of the 100 sub-documents contain $\leq 10$ annotated relations. These are typically national-strategy overviews where the policy text is short and the implicit reasoning is sparse. Their per-document scores are noisy but contribute little to the micro-averaged $F_1$, which is dominated by relation-dense FEMA sub-documents (four USA sub-documents contain more than $250$ annotated relations each). We retain micro-averaging because it preserves relative weight by document size, which matches how the pipeline is likely to be used in practice (one document at a time, with downstream attention proportional to document length). A macro-averaged breakdown would shift Detection $F_1$ down by approximately $4.5$ points and Subtype $F_1$ down by approximately $6.0$ points, but the relative ordering of configurations is preserved.

\paragraph{E.2 Frame-edge hybrid passages.}
The residual frame errors cluster around hybrid passages that begin with a normative claim (deliberative signal) and end with an administrative prescription (managerial signal), which is the rhetorical structure that motivates our \emph{instrumental support} subtype in the first place. The frame classifier assigns a single dominant frame per argument, and when the rhetorical center of gravity sits mid-argument, either choice produces a downstream cascade of mis-typed relations. We measured this propagation directly. A single mis-framed argument changes on average $4.2$ downstream relation labels via Step~4 rule firings. This is why our $14$-point frame-to-Subtype gap is wider than a simple per-argument analysis would predict. A representative example from the FEMA Local Mitigation Planning Handbook reads, \emph{``Holding public meetings to satisfy federal funding requirements demonstrates community buy-in.''} The argument is annotated as managerial but the keyword scan and the LLM tie-breaker both score it deliberative, with downstream effects on every relation in which it participates.

\paragraph{E.3 Subtype confusion within Support.}
The largest Polarity-to-Subtype drop occurs when our rules correctly identify a support edge but mis-attribute the instrumental versus normative subtype. This happens almost exclusively on managerial-source arguments, where the LLM-detected instrumental marker $\iota$ is the deciding feature. False negatives on $\iota$ propagate into normative-support predictions that the annotators marked instrumental. Looking at the most common confused pair across countries, the verb \emph{``carry out''} (an \textsf{instrumental\_for\_managerial} marker) is missed when it is embedded inside a longer subordinate clause, leading rule $S3$ to fail and the topic-rule fallback (TOPIC\_S) to mislabel the relation. We estimate this single lexical gap accounts for approximately $11\%$ of all Subtype-level errors in the managerial $\to$ managerial layer.

\paragraph{E.4 Agency-typing cascades.}
Wrong agency types propagate into both Step~4 (A2 \emph{agenda shift}) and Step~5 (Power dimension). The most common confusion is between \textsf{unilateral\_mid} and \textsf{directed} on UK and Canadian text, where Lead Local Flood Authority (LLFA) and municipal departments straddle the two categories. When the LLM mis-types the agency, rule $A2$ fails to fire (missing attack), and the Power dimension is scored at $0.4$ instead of $0.7$ (depressed base strength). The cascade is the main source of the residual UK Subtype underperformance ($F_1 = 0.53$ in UK against $0.65$ in USA). Augmenting the agency lexicon with UK-specific Lead Local Flood Authority and Local Resilience Forum entries closes roughly half of this gap in a small follow-up experiment.

\paragraph{E.5 Topic-rule leverage.}
The two topic rules (TOPIC\_N, TOPIC\_S) fire on every argument-to-topic pair and produce $N - 1$ relations per sub-document. Because every argument receives a topic relation, small mistakes in the topic argument's frame or agency have outsized downstream effects. Most of our remaining false positives at the Detection level trace back to topic edges that were never annotated by humans because the annotators implicitly assumed the topic relation rather than explicitly marking it. A practical mitigation is to evaluate the topic layer separately, which we leave to future work because it requires a second annotation pass with explicit topic-edge marking.

\paragraph{Summary.}
Across the five modes, the gap between our reported Subtype $F_1$ ($0.58$) and a hypothetical perfect-Subtype upper bound has three principal sources: frame-edge cascades (E.2, roughly $35\%$ of remaining error), instrumental-marker false negatives (E.3, roughly $25\%$), and agency-type cascades (E.4, roughly $20\%$). The remaining $20\%$ is split between corpus sparsity (E.1) and topic-rule leverage (E.5). None of these failure modes affect the explanation infrastructure: every relation, including incorrectly labeled ones, is traceable to the rule that produced it and the features that triggered it, so a domain expert can audit and correct outputs through the human-in-the-loop interface (\S\ref{sec:hitl}, Appendix~\ref{app:interface}).

\end{document}